\pdfoutput=1
\documentclass[runningheads]{llncs}
\usepackage[T1]{fontenc}

\usepackage{newtxtext}
\usepackage[varvw,varbb]{newtxmath}
\usepackage[mathcal]{euscript}

\usepackage{graphicx}
\usepackage{tikz}\usetikzlibrary{positioning,arrows.meta}

\newcommand{\Jeff}{\mathcal{J}}
\newcommand{\Source}{\mathcal{D}_{\text{train}}}
\usepackage{hyperref}

\makeatletter
\@ifundefined{credits}{\newenvironment{credits}{\par\addvspace{17\p@}\small}{\par\addvspace{\bigskipamount}}}{}
\providecommand{\ackname}{Acknowledgements}

\makeatother

\begin{document}
%
\title{Flagging Batch Shift in Flow Cytometry using Jeffreys Divergence between Gaussian Mixtures}
%
\titlerunning{Flagging batch shift with Jeffreys divergence}
%

\author{Jonathan Legrand\inst{1} \and
Aguirre Mimoun\inst{2} \and
Audrey Bidet\inst{3} \and
Pierre-Yves Dumas\inst{4} \and
Christèle Etchegaray\inst{1}$^{\dagger}$\and
Baudouin Denis de Senneville\inst{1}$^{\dagger}$
}
\authorrunning{J. Legrand et al.}
\institute{Univ. Bordeaux, CNRS, Inria, Bordeaux INP, IMB, UMR 5251,
F-33400 Talence, France\\
\email{jonathan.legrand@math.u-bordeaux.fr} \and
CHU Bordeaux, Laboratoire d'Hématologie, F-33000 Bordeaux, France \and
Laboratoire d'Hématologie Biologique, CHU Bordeaux, Avenue Magellan,
33600 Pessac, France \and
Univ. Bordeaux, INSERM, BRIC, U1312, CHU Bordeaux, Service d'Hématologie
Clinique et de Thérapie Cellulaire, F-33000 Bordeaux, France \\
$^{\dagger}$ These authors jointly supervised this work.
}

\maketitle              
\begin{abstract}

Machine learning models for clinical flow cytometry are usually
evaluated on data drawn from their training distribution. 
In routine practice, however, acquisition conditions and reagents may change,
potentially making predictions unreliable without warning. 
We propose a model agnostic criterion for prediction reliability
based on a notion of distance from the training distribution:
each patient sample
is represented by a
Gaussian mixture model, and its Jeffreys divergence
to a reference fitted on the training data is estimated
using Monte Carlo integration.
We illustrate our approach on real-life shifted cohorts of
acute myeloid leukaemia patients, using a classifier trained
to predict \textit{NPM1} mutation status.
First, we show graphically and statistically that pairwise divergences capture acquisition cohort structure.
We then calibrate a rejection threshold on a subset
of shifted cohort samples and evaluate it on held-out subjects.
Across 20 runs, the misclassification rate of rejected samples was 2.5 times larger than for retained samples. 
These results suggest that input
divergence can help identify unreliable predictions,
providing a strong candidate component for abstention frameworks,
along with quality control measures and classifier confidence.

\end{abstract}

\section{Introduction}

Flow cytometry measures tens of fluorescence and scatter parameters, for thousands of single cells and is central to the diagnosis of haematological malignancies and immune disorders. Computational methods, from automated gating to supervised classification and deep learning, increasingly aim to assist or automate its interpretation~\cite{hu_application_2022}. Most reported performances are, however, obtained within the distribution used for training~\cite{de_marchi_putting_2026}. External or multicentre validation is comparatively rare, although acquisition-related biases across time, laboratories and instruments are well documented \cite{maecker_model_2010}.

Several batch correction methods have been developed to mitigate technical variability across cytometry datasets (see~\cite{nunez-nepomuceno_survey_2025}, Table III, for a comprehensive review). However, when the input distribution deviates from the training distribution, a classifier can keep producing confident outputs, so errors are silent~\cite{ovadia_can_2019}. Selective classification, in which a model may abstain on inputs it is likely to misclassify, is a classical answer to this problem~\cite{el-yaniv_foundations_2010}. Abstention rules based on the classifier's own confidence, however, are known to degrade under shift, which motivates scores computed from the input distribution, independent of the classifier~\cite{ovadia_can_2019}.


In parallel, the use of probability distributions to represent entire cytometry samples,
and of divergence measures to quantify differences between them, has been investigated.
Early work proposed clustering on embed
approximations of Fisher information distance~\cite{carter_dimensionality_2008},
which was further applied
to samples from acute lymphoblastic leukaemia patients~\cite{finn2009}.
Transport-based distances have also been used to compare cytometry samples and build robust gating techniques~\cite{del_barrio_optimalflow_2020}. Outside cytometry, GloScope summarises each single-cell RNA-seq sample with a Gaussian mixture, estimates the symmetrised KL divergence between samples by Monte Carlo integration, and uses the resulting distance matrix for sample-level visualisation and quality control, including batch effects~\cite{wang2024visualizing}. In all these works the divergence serves to represent, compare and cluster samples.

Here we address the following, complementary question: can we quantify how far a new sample is from the data a classifier was trained on, cheaply and in a way that is informative about the classifier’s error rate? We build on the sample-as-mixture and Monte Carlo divergence estimators of Wang et al. \cite{wang2024visualizing} and make three contributions:
\begin{enumerate}
\item We use Jeffreys divergence between sample-level Gaussian mixtures to map the batch structure of a collection of acquisitions.
\item We define an abstention rule placed upstream of a frozen classifier, with a threshold calibrated on a held-out subset of the test cohort.
\item We show that the divergence to the training distribution is informative about
the error rate of a frozen classifier, on a task of \textit{NPM1} mutation status
prediction from flow cytometry.
\end{enumerate}

\section{Methods}
\subsection{Problem Setting}

Let $X^n = \{x^n_1, \dots, x^n_{C^n}\}$ be a set of $C^n$ cells sampled from patient $n$,
where each cell $x^n_c \in \mathcal{X} = \mathbb{R}^D$ is
described by $D$ cellular features. In a multiple instance learning framework, each cell is an instance, while $X^n$ is a bag.
Let $\mathbb{N}^\mathcal{X}$ denote the space of bags over the instance space $\mathcal{X}$.
A multiple instance classifier $f: \mathbb{N}^\mathcal{X} \rightarrow \mathcal{Y}$ is trained once on a fixed
dataset with corresponding sample-level labels $\Source = \{(X^1, y^1), \dots, (X^N, y^N)\}$ so as to produce predictions for unseen samples
$\hat{y}^m = f(X^m)$.

\begin{figure}[t]
\centering
\begin{tikzpicture}[node distance=4mm,every node/.style={font=\scriptsize},blk/.style={draw,rounded corners,align=center,minimum height=7mm,inner sep=2pt}]
\node[blk] (a) {Sample\\cells $X^m$};
\node[blk,right=of a] (b) {Fit GMM $\mu^m$};
\node[blk,right=of b] (c) {Estimate Jeffreys divergence to $\mu^0$\\ $d_m = \hat{\Jeff}(\mu^0,\mu^m)$};
\node[blk,right=of c] (d) {$d_m>\tau$?};
\node[blk,above right=2mm and 4mm of d] (e) {Abstain:\\expert review};
\node[blk,below right=2mm and 4mm of d] (f) {Frozen\\classifier output $f(X^m)$};
\draw[-{Latex}] (a)--(b);\draw[-{Latex}] (b)--(c);\draw[-{Latex}] (c)--(d);
\draw[-{Latex}] (d)--node[above,sloped]{yes}(e);\draw[-{Latex}] (d)--node[below,sloped]{no}(f);
\end{tikzpicture}
\caption{Abstention module at evaluation time, after calibration. The training
set is summarised by the reference density $p^0$, and the calibration on
labelled target samples yields the threshold $\tau$. For a new sample, the
score $d_m$ is the Monte Carlo Jeffreys divergence between its density $p^m$
and $p^0$. It depends only on the input density and not on the classifier.
The prediction of the frozen classifier $f$ is withheld if $d_m>\tau$.}
\label{fig:framework}
\end{figure}
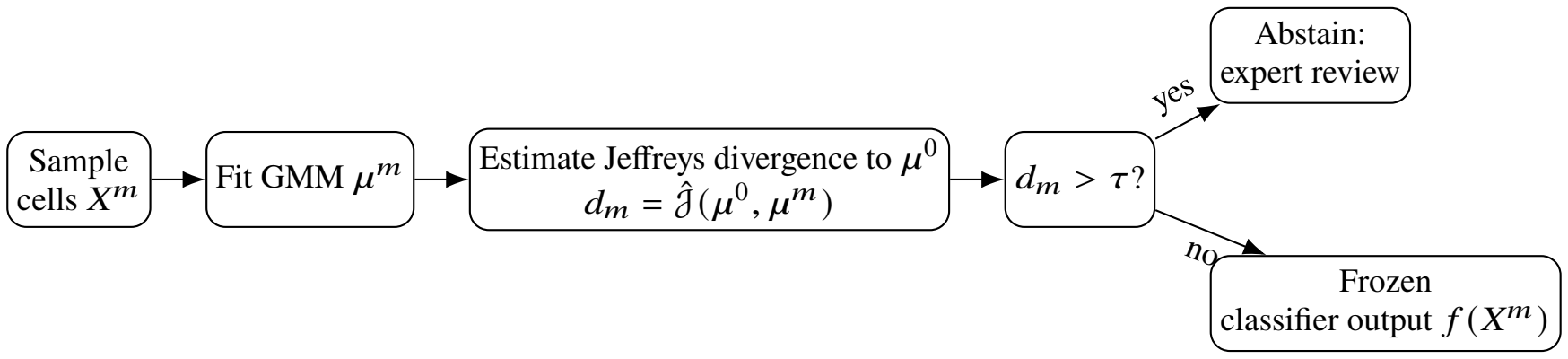

An abstention module computes a scalar score $d_m$
from $X^m$ alone, measuring how far the sample lies from the training
distribution, and withholds the prediction of $f$ when $d_m > \tau$
(Fig.~\ref{fig:framework}). The threshold $\tau$ is calibrated on a small
labelled subset of the target data (Section~\ref{sec:setup-rejection}).

\subsection{Data}
\subsubsection{Task.} We consider the prediction of \textit{NPM1} mutation status
(mutated vs.\ wild-type) in acute myeloid leukaemia from bone marrow flow
cytometry samples. The \textit{NPM1} status of a patient informs early treatment decisions, however, molecular testing can take up to several weeks. Thus, recent studies focused on reliably predicting molecular features
of acute myeloid leukaemia from flow cytometry, which is available early in the patient care~\cite{legrand_interpretable_2026,lewis_automated_2024}.

\subsubsection{Cohorts.}
The training set $\mathcal{D}_{\mathrm{train}}$ contains 172 patients
diagnosed with acute myeloid leukaemia, treated at the Bordeaux
University Hospital Center between 2019 and 2020 and with known
\textit{NPM1} mutation status. These patients underwent bone
marrow aspirates which were stained and analysed following
the recommendations
of the GEIL immunophenotyping panel~\cite{publication_panel_2018}.

On the other hand, the test set is constituted of two separate cohorts.
The first, referred to as the \textit{Time-shifted} dataset, comprises 156 patients analysed
using the source dataset protocol but acquired between 2013 and 2018, 
thereby introducing a temporal shift. 
The second, referred to as the \textit{Panel-shifted} dataset,
comprises 242 patients analysed through the ELN protocol~\cite{heuser_2021_2021}
and uses different fluorochromes.
As the GEIL and ELN panels only partially overlap, all analyses were
restricted to the seven shared features: forward scatter,
side scatter, and the surface markers CD13, CD33, CD34, CD45, and CD7.
The Time-shifted cohort mimics the drift of a single laboratory over time,
whereas the Panel-shifted cohort mimics a change of reagents and acquisition
protocol, thereby providing a suitable test-case for our method.

\begin{table}[ht]
\caption{Cohort characteristics. WT: wild-type. Percentages are relative to
the cohort size.}
\label{tab:cohorts}
\centering
\begin{tabular}{lccc}
\hline
Dataset &$\quad$  $N$ $\quad$  & $\quad$ \textit{NPM1}-WT, $n$ (\%) $\quad$  &$\quad$  \textit{NPM1}-mutated, $n$ (\%)\\
\hline
Source ($\Source$) & 172 & 120 (69.8\%) & 52 (30.2\%) \\
Time-shifted       & 156 & 108 (69.2\%) & 48 (30.8\%) \\
Panel-shifted      & 242 & 173 (71.5\%) & 69 (28.5\%) \\
\hline
\end{tabular}
\end{table}

The proportion of \textit{NPM1}-mutated patients did not differ significantly
across cohorts (Pearson $\chi^2$ test on the $3\times2$ contingency table,
$\chi^2(2) = 0.27$, $P = 0.87$; Table~\ref{tab:cohorts}),
so label shift is unlikely to account for performance
differences between cohort.

The studies complied with the Declaration of Helsinki and were approved by
French authorities (CNIL N°915\,285, CCTIRS N°15.319). Additional information
regarding data acquisition
and demographics can be found in \cite{legrand_interpretable_2026}.

\subsection{Preprocessing}

Preprocessing consisted of three sequential steps: (i) removal of events detected at the signal acquisition margins; (ii) fluorescence spillover compensation using the compensation matrix embedded in each data file; and (iii) transformation of fluorescence intensities using the $\operatorname{arcsinh}(x/5)$ function. Quality-control plots were generated for each sample to assess margin removal. These plots were archived and manually inspected to verify the integrity of the preprocessing procedure. Each subject was represented by a subsample of 1000 cells randomly drawn without replacement from its FCS file (i.e. $\forall n \in \{1, \dots, N\}, C^n = C = 1000$).
Marker-wise Z-score normalisation was performed by subtracting the mean and dividing by the standard deviation, with both statistics estimated from $\mathcal{D}_{\mathrm{train}}$.

\subsection{Cell density estimation with Gaussian Mixture Models}

To provide a compact representation of high-dimensional cytometry data, each
sample is modelled using a Gaussian Mixture Model (GMM), defined as a convex combination of $K$
multivariate Gaussian distributions:
\begin{equation}
\mu = \sum_{k=1}^{K} \pi_k \mu_k, \quad \mu_k = \mathcal{N}(m_k,\Sigma_k),
\end{equation}
where $\pi_k\ge 0$ denotes the mixing weight of component $k$, subject to
$\sum_k\pi_k=1$. The parameters $m_k$ and $\Sigma_k$ represent the mean vector
and covariance matrix of the corresponding Gaussian component, respectively.
The mixture parameters are estimated using the expectation-maximisation
algorithm \cite{deisenroth_mathematics_2020}.

\subsection{Jeffreys divergence between mixtures}

Let $\mu^i$ and $\mu^j$ be two GMM distributions with respective densities $p^i$ and $p^j$.
To quantify the dissimilarity between cytometry samples, we used Jeffreys
divergence, defined as the symmetrized Kullback--Leibler divergence:
\begin{align}
\Jeff(\mu^i,\mu^j) &:= D_{KL}(\mu^i\|\mu^j) + D_{KL}(\mu^j\|\mu^i) \nonumber\\
&= \int_{\mathbb{R}^D}\big(p^i(z)-p^j(z)\big)\log\frac{p^i(z)}{p^j(z)}\,\textrm{d}z
\end{align}
Although the Jeffreys divergence is finite for Gaussian mixture distributions,
it generally lacks a closed-form expression, unlike the case of single Gaussian
distributions. We therefore estimated it using Monte Carlo integration.
Since Gaussian mixtures can be sampled exactly and efficiently, we
independently draw $S$ samples $z_1,\ldots,z_S\sim\mu^i$ and $S$ samples
$z'_1,\ldots,z'_S\sim\mu^j$.
The Monte Carlo estimator is given by:
\begin{equation}
\hat{\Jeff}(\mu^i,\mu^j)=\frac{1}{S}\sum_{s=1}^{S}\Big[\log\frac{p^i(z_s)}{p^j(z_s)}
+\log\frac{p^j(z'_s)}{p^i(z'_s)}\Big].
\label{eq:mc}
\end{equation}
This estimator is unbiased, with variance decreasing proportionally to $1/S$~\cite{hershey_approximating_2007}.

\subsection{Visualization of inter-cohort divergence}
\label{sec:setup-pairwise}

To assess whether the Jeffreys divergence captures cohort structure, we computed pairwise divergences
between all subjects across the cohorts included in the study,
yielding a symmetric pairwise dissimilarity matrix. 
We used $K=8$ full-covariance components as default parameters for the GMM.

To visualize the resulting dissimilarity structure, we projected the matrix into a two-dimensional space using multidimensional scaling (MDS)~\cite{borg_modern_2013}.
Because the Jeffreys divergence is not a metric, as it does not generally satisfy the triangle inequality, we used non-metric MDS, which seeks to preserve the rank order of the pairwise dissimilarities. Its computation was initialized using classical MDS.

To quantify the extent to which the observed dissimilarity structure was associated with cohort membership, we performed a permutational multivariate analysis of variance (PERMANOVA) on the full pairwise dissimilarity matrix, using cohort identity as the grouping factor and $B=1000$ permutations (see Section~\ref{app:permanova}). We report the associated $P$-value and the coefficient of determination ($R^2$), which quantifies the proportion of variation in the dissimilarity structure attributable to cohort membership.

\subsection{Abstention framework}
\label{sec:setup-rejection}

We used a threshold-based rejection criterion based on the Jeffreys divergence estimation to
identify samples for which the classifier's predictions were likely to be unreliable. 

\subsubsection{Training set representation.}
The reference mixture $\mu^0$ is the GMM underlying the Fisher kernel embedding
of the classifier (Section~\ref{sec:experimental-setup}). It is fitted once on the
cells pooled from $\Source$, and is kept fixed across all GMM configurations
considered below.

\subsubsection{Test subjects representation.}
Each subject of the test set was represented by a fitted GMM from the same configuration.
Contrary to the visualization analysis, the
subject-level GMM configuration was here selected on a calibration subset.
We considered six GMM configurations,
differing by the number of components $K \in \{1, 4, 8\}$
and by the structure imposed on the covariance matrices (full or diagonal).
The test subjects were randomly separated into a
calibration and an evaluation subset using a $20/80$ stratified split.

\subsubsection{Dissimilarity to the training set.}
The dissimilarity of each calibration subject $\ell$ from this reference
distribution was computed as $d_\ell = \hat{\Jeff}(\mu^0,\mu^\ell)$. The obtained
values, sorted in decreasing order, defined successive
rejection thresholds $\{\tau_\ell\}_{1\leq \ell \leq L}$.
Then, for each threshold $\tau_\ell$, the proportion $r_\ell$ of
below-threshold (i.e. retained) subjects was computed, and the set
$\{(\tau_\ell,r_\ell)\}_{1\leq \ell \leq L}$ defined a
so-called retention curve. Following classification of retained subjects of the calibration set, distinct retention curves were obtained for correct and incorrect classification outputs, with proportions denoted by $r_\ell^{\pm}$, respectively.

\subsubsection{Selection of test GMM configuration.}
The selection of final GMM configuration for test subjects was based on its ability to realize a trade-off between retaining correctly classified subjects while rejecting misclassified ones. To this end, we computed the mean retention gap (MRG),
defined as the signed difference between the two retention curves, averaged across thresholds.
It is given by
\begin{equation}
\text{MRG} := \frac{1}{L}\sum_{\ell=1}^{L}(r_\ell^{+} - r_\ell^{-}).
\label{eq:abst_score}
\end{equation}
Hence, a positive gap indicates that misclassified subjects are rejected more rapidly than correctly classified subjects. The GMM configuration with the largest mean signed gap was selected.

\subsubsection{Threshold optimization.}
Using the calibration subset and the corresponding GMM configuration,
an optimal threshold $\tau$ was defined as the threshold for which the retention gap was maximal:
\begin{equation}
\tau := \text{argmax}_{1\leq \ell \leq L}\{r_\ell^{+} - r_\ell^{-}\}\,.
\end{equation}
    
\subsubsection{Evaluation.}
The effectiveness of the selected abstention framework (GMM configuration and threshold)
was assessed on the evaluation subset by comparing misclassification rates between
rejected and non-rejected samples.

\subsection{Experimental setup}
\label{sec:experimental-setup}
We used a support vector machine classifier with a Laplacian kernel on Fisher embeddings,
following Perronnin et al. \cite{perronnin_improving_2010}.
The Fisher embedding relies on its own GMM, fitted on the cells pooled from $\Source$.
We therefore use this model as the reference distribution $\mu^0$. Throughout the
experiments, $\mu^0$ parameters are set to $K=5$ components with diagonal covariance matrix.

Results are reported as the mean $\pm$ standard deviation over 20 independent runs with random seeds variations to account for variability in cell sampling, GMM fitting, and partitioning of the test set into calibration and evaluation subsets for the abstention framework.
As the calibration subset is redrawn at each run, the selected threshold $\tau$ also varies across runs. All performance measures (e.g. rejection and misclassification rates) were computed on the evaluation subset using the threshold selected in the same run, and then averaged.
The Monte Carlo approximation of the Jeffreys divergence used $S=256$ samples in all runs.
The scripts used for divergence calculation, plots, statistical analysis, and our implementation of the Fisher embedding, are all available publicly at \url{https://github.com/jonathan-legrand/jeffreys-flow-cytometry}

\section{Results}

\subsection{Visualization of inter-cohort divergence and batch structure}

\begin{figure}
    \centering
    \includegraphics[width=0.7\textwidth]{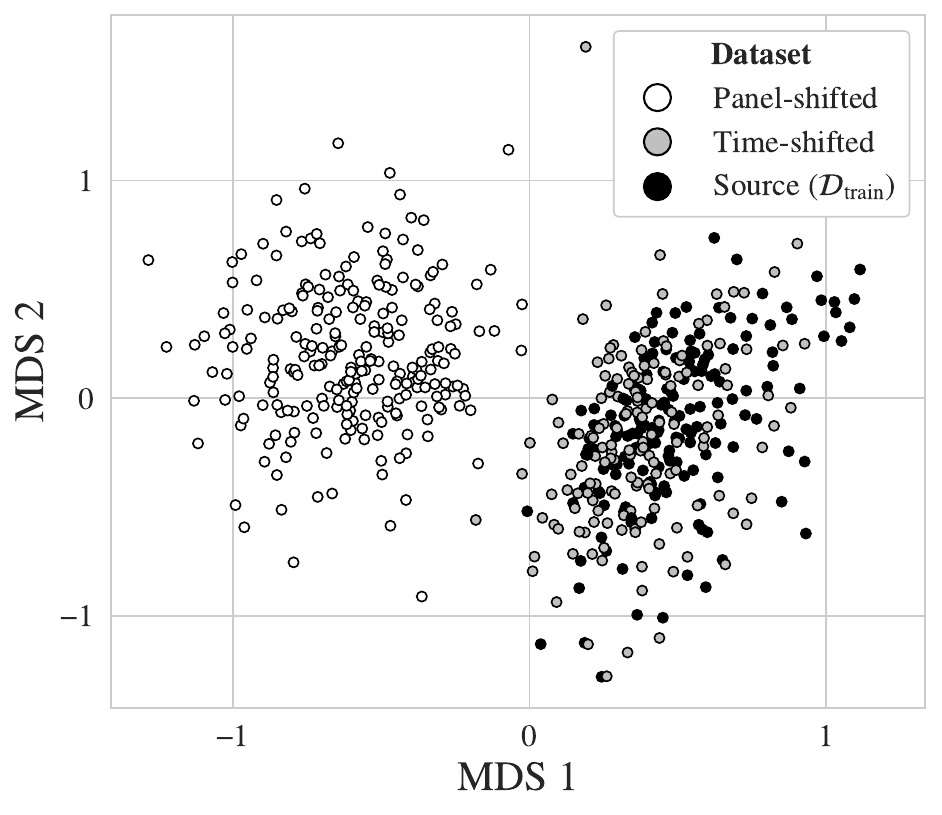}
    \caption{Non-metric MDS embedding of pairwise Jeffreys divergences between subject-level GMMs. Each point represents a patient, coloured according to its acquisition protocol.}
    \label{fig:mds}
\end{figure}

Figure~\ref{fig:mds} shows a two-dimensional MDS embedding of the pairwise estimated Jeffreys divergences between subjects. 
Samples from the Panel-shifted protocol (white) cluster on the left side
of the plane, while the Time-shifted (grey) and $\Source$ (black) cohorts are overlayed on
the right hand side. These panel-induced clusters are very distinguishable along the first MDS axis, with few points on the boundary. On the other hand, the Time-shifted and Source datasets seem to share the same distribution.
A PERMANOVA confirmed a strong association between cohort membership and the divergence structure, with cohort identity accounting for 61\% of the observed variation ($P=10^{-3}, R^2=0.61$). These results indicate that the proposed divergence captures batch-related variation across the three cohorts.

\subsection{Calibration and abstention under shift}
\label{sec:batch-effects}
Before rejection, the classifier achieved a mean accuracy of 
81\% (SD 2\%) across the two test cohorts.
Accuracy was 84\% (SD 1\%) on the Time-shifted
cohort and 79\% (SD 2\%) on the Panel-shifted cohort.
Across 20 independent runs, the best GMM configuration on average
used a single full-covariance Gaussian component.

\begin{figure}
\centering
\includegraphics[width=0.99\textwidth]{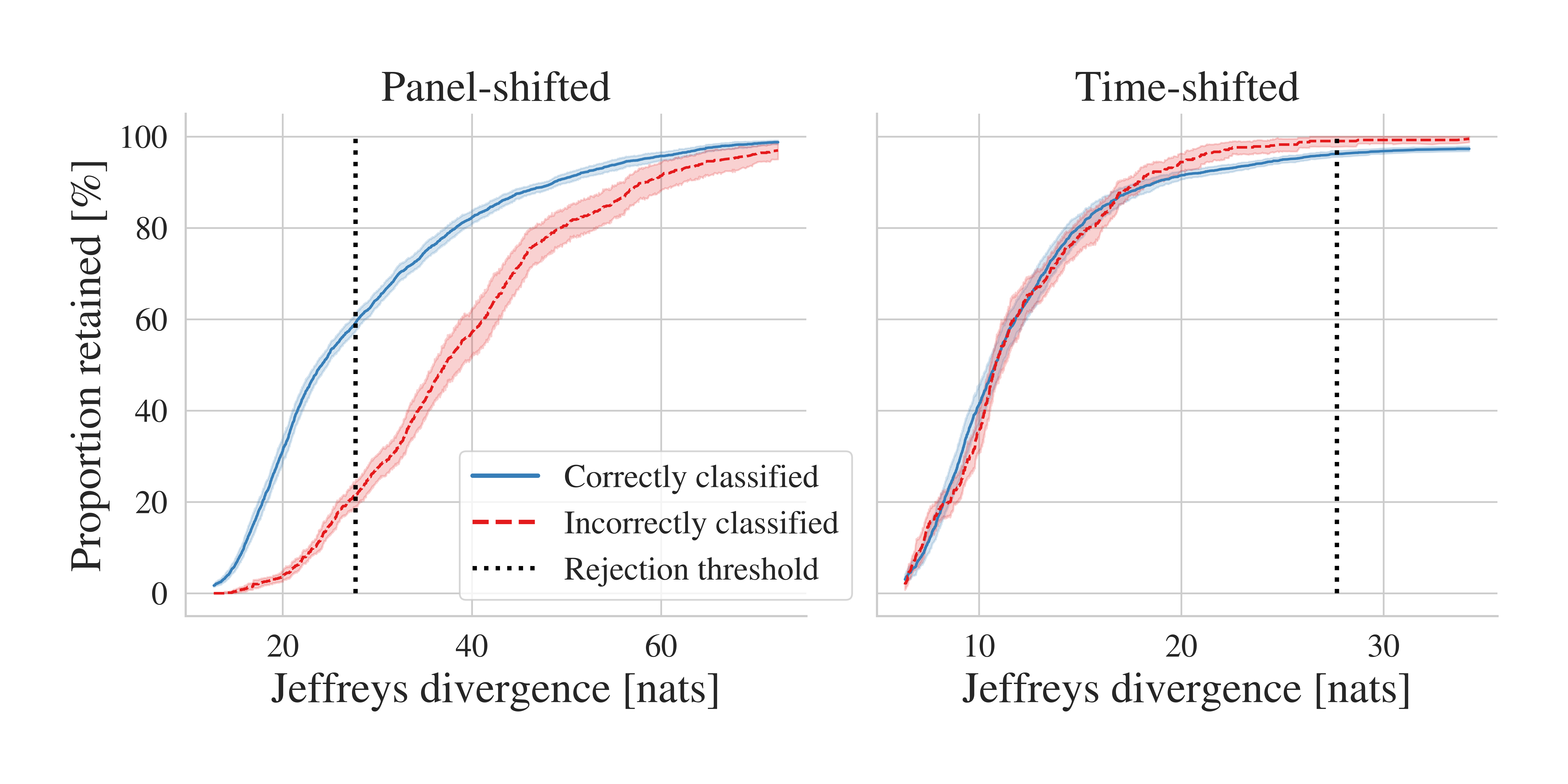}
\caption{Mean retention curves on the evaluation subsets across 20 independent runs for the selected abstention GMM configuration. The curves show the proportions of correctly (blue, plain) and incorrectly (red, dashed) classified subjects retained (MRG) as the rejection threshold varies. Shaded regions represent the 95\% confidence intervals across 20 independent runs, and the dotted vertical line marks the median of the 20 optimal thresholds for that configuration.} \label{fig:rejection-curves}
\end{figure}

Figure~\ref{fig:rejection-curves} shows the resulting retention curves on each test
cohort separately. On the Panel-shifted cohort (left), the curve of correctly classified
subjects lies above that of misclassified subjects over almost the entire
range of thresholds, meaning that misclassified subjects have a larger
dissimilarity with the training set. At the median selected threshold,
about 60\% of correctly classified subjects were
retained, against about 20\% of misclassified subjects.

On the Time-shifted cohort (right), the two curves largely overlap: they are
indistinguishable up to about 17 nats, and the misclassified subjects are
retained slightly more than the correctly classified ones at higher
divergences. 
The selected threshold lies in the upper
tail of the distribution of divergences, and retains more than 95\% of both
correctly and incorrectly classified subjects, so that almost no
Time-shifted subject is rejected.

Pooling both test cohorts together, at the calibrated threshold,
the selected configuration rejected an
average of 30.9\%
of evaluated subjects (SD 2.7\%).
Analysis of rejected subjects labels showed that they had a
misclassification rate of 33\% (SD 1.8\%),
while the retained subjects only reached 13\% (SD 5.4\%), approximately a 2.5 fold improvement.
The mean rejection gap varied little across the
six tested GMM configurations (Fig.~A\ref{figsup:methods-evaluation}).

\section{Discussion}




We proposed to flag unreliable predictions of a frozen classifier, using the
divergence between the distribution of a new sample and that of the training
distribution. Our approach was illustrated on a challenging multi-cohort dataset
combining a temporal shift and a panel shift.
The pairwise divergences exposed the acquisition
structure of the data:
the Panel-shifted cohort lay clearly apart from the
training cohort, whereas the Time-shifted cohort remained close to it.
This is consistent with previous findings on MDS embedding of symmetric KL divergence
in single cell data~\cite{del_barrio_optimalflow_2020}.
Moreover, a single threshold on the divergence to the
training reference flagged a substantial share of
misclassified samples while retaining most correctly classified ones,
making Jeffreys divergence a promising candidate for
abstention modules in a medical context. Our score has two practical
advantages: it is independent of the classifier, so it can accompany any frozen
model, and it is cheap, as one GMM fit and one Monte Carlo estimate per sample
suffice.

The number of components and the covariance structure of the subject-level GMMs
was tuned on the  calibration subset to maximise the mean retention gap.
The configuration selected on the calibration subset was a single full covariance Gaussian,
meaning that each patient was summarised by a mean vector and a covariance matrix only.
This result is consistent with the observation that
change of fluorochromes or instrument settings mostly
act as translations and
rescalings of marker intensities, which are global first 
and second order effects.
Subsequent analysis showed that the choice of configuration had
little influence on the rejection performance (Fig.~A\ref{figsup:methods-evaluation})
and on the pairwise divergence analysis (Fig.~A\ref{figsup:mds-k-1}).
However, this may not hold for finer domain shifts
where richer mixtures could be needed.

The threshold was set here to maximise the gap between the two retention
curves. Other choices are possible: one may require a target prediction quality
on the retained samples or, conversely, cap the rejection rate so as not to
overload human reviewers. At any rate, as the blue curve of the
Panel-shifted cohort
Fig.~\ref{fig:rejection-curves}
is consistently above the red one, we expect our method to give 
useful results on a wide range of thresholds.

When looking at performance cohort-wise,
it appeared that the benefit of the rejection rule came mostly
from the Panel-shifted cohort.
This is consistent with the fact that panel shifted samples
were clustered away from the training distribution
(Fig.~{\ref{fig:mds}}) and that they were harder to
classify for our model
(Section~\ref{sec:batch-effects}).
On the other hand, for the Time-shifted cohort, the score was not informative and almost no
prediction was withheld. Its subjects were close to the training data, so the
errors made there were not likely related to the divergence of the sample from the
training distribution.
Thus, the dissimilarity should not be read as a stand-alone decision rule. It measures
input shifts which are important enough to affect the density and divergence estimates,
but that
does not cover all types of possible classification errors.
We rather view the divergence as a promising candidate component of a larger
multi-parameter abstention system at the output of a classifier, combining also
quality control indicators and classifier confidence~\cite{el-yaniv_foundations_2010}.

\subsubsection{Limitations and future work.} Our evaluation is limited in scope:
it relies on a single task, on a monocentric cohort, and on a single classifier.
Validation on
other datasets and on multicentric cohorts is therefore required, ideally
in a prospective setting that accounts for the cost of expert review.

The behaviour of the score may also depend on the type of classifier, and it
remains to be determined whether some configurations are better served by this
divergence than others. Likewise, we did not benchmark alternative density
models (e.g. non-parametric estimates or other generative models) or
alternative dissimilarities (e.g. Wasserstein distances~\cite{del_barrio_optimalflow_2020},
maximum mean
discrepancy). Alternatives for the reference built from the
training distribution (pooled mixture, per-batch, nearest-batch,
per-class references) could also be explored.

\section{Conclusion}

We introduced a classifier-agnostic abstention module for flow cytometry based on
GMM cell densities and a Monte Carlo Jeffreys divergence, a divergence visualization
that exposes the batch structure, and an evaluation protocol with real life shifted datasets.
Across 20 independent runs, the rule rejected approximately 31\% of evaluated subjects,
and rejected subjects had a misclassification rate 2.5 times higher than
retained subjects.
These results support the use of statistical divergences as building
blocks for multi-parameter abstention systems at the output of a classifier.

\begin{credits}
\subsubsection{\ackname}
The CytoFLAM project has received the joint financial support of the Inria
Centre at the University of Bordeaux and of the Nouvelle-Aquitaine Region in
the scope of their ``Research and Platform'' call. This project was partially
funded by the AOI of the Bordeaux University Hospital. We would like to thank
the clinical data management unit of Toulouse University Hospital.
During the preparation of this manuscript, the authors used Le Chat by Mistral AI, Claude by Anthropic and ChatGPT by OpenAI to refine the clarity and language of the text. After using this tool, the authors carefully reviewed and edited the content and take full responsibility for the final version.

\subsubsection{Author contributions.}
JL: development of methodology and software, formal analysis, writing original draft.
AM: conceptualization of the study, data curation, investigation.
AB: conducted the data curation. 
PYD: conceptualization of the study. 
CE: conceptualization of the study, funding acquisition, supervision, reviewing and editing the draft. 
BDS: conceptualization of the study, funding acquisition, supervision, reviewing and editing the draft.

\subsubsection{Data availability statement.}
The datasets analysed during the current study are not publicly available due
to privacy or ethical restrictions but are available from the corresponding
author on reasonable request.

\subsubsection{Ethics approval and consent to participate.}
This study was conducted in accordance with the Declaration of Helsinki. The
cohorts analysed in this study were collected as part of the ICAML study
(ClinicalTrials.gov identifier NCT03449745) and the DATAML study. The DATAML
study was approved by the relevant French regulatory authorities: the
Commission Nationale de l'Informatique et des Libert\'es (CNIL, reference
N\textdegree915\,285) and the Comit\'e Consultatif sur le Traitement de
l'Information en mati\`ere de Recherche dans le domaine de la Sant\'e (CCTIRS,
reference N\textdegree15.319). Informed consent to participate was obtained
from all patients included in the study.
\end{credits}

\appendix
\newpage

\section{Appendix}
\subsection{PERMANOVA of the dissimilarity matrices}
\label{app:permanova}

To test whether the divergences between subjects carry information about the
acquisition group (Source cohort, Time-shifted cohort, Panel-shifted cohort), we used
permutational multivariate analysis of variance
(PERMANOVA)~\cite{anderson_permutational_2017}. PERMANOVA applies to any
dissimilarity matrix and does not require the distances to be Euclidean or
the data to be normally distributed, which matters here since the Jeffreys divergence is not a Euclidean distance.

Let $d_{ij}:=\hat{\Jeff}(\mu^i,\mu^j)$ be the distance between subjects $i$ and $j$, $N$ the number of
subjects, $g$ the number of groups, and $n_l$ the size of group $l$. The total
and within-group sums of squares were computed directly from the pairwise
divergences:
\begin{equation*}
SS_T = \frac{1}{N}\sum_{i<j} d_{ij}^2, \qquad
SS_W = \sum_{l=1}^{g} \frac{1}{n_l} \sum_{\substack{i<j\\ g_i=g_j=l}} d_{ij}^2,
\end{equation*}
where $g_i$ is the group of subject $i$. The pseudo-$F$ statistic is given by
\begin{equation*}
F = \frac{(SS_T - SS_W)/(g-1)}{SS_W/(N-g)}.
\end{equation*}
Large values indicate that subjects of the same group are closer to each
other than expected by chance. When the distances are Euclidean, $F$ equals
the classical ANOVA $F$ ratio.

Since the sampling distribution of $F$ is unknown for general
dissimilarities, significance was assessed using a permutation testing procedure.
Group labels were
randomly reassigned to subjects $B=1000$ times, keeping the dissimilarity matrix
and group sizes fixed, and a null statistic $F_b$ was recomputed at each iteration.
The
$P$-value was computed as:
\begin{equation*}
P = \frac{1 + \sum_{b=1}^{B} \mathbf{1}\left\{F_b \geq F_{obs}\right\}}{1 + B}\,.
\end{equation*}
Note that the smallest attainable value is $1/1001$.
The $R^2$ score, which indicates the proportion of total variation due to the groups effect,
was also reported along with the $P$-value:
\begin{equation*}
    R^2 = 1 - \frac{SS_W}{SS_T}
\end{equation*}

\newpage
\subsection{Influence of GMM parameters on mean retention gap}
\begin{figure}[ht]
    \centering
    \includegraphics[width=0.7\linewidth]{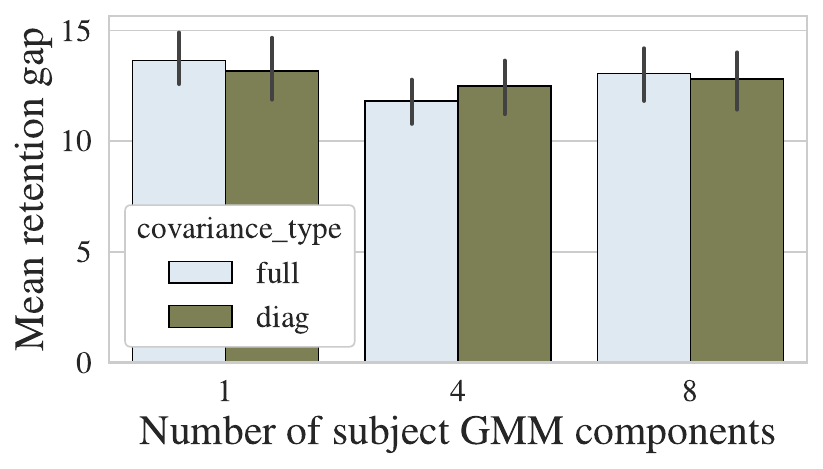}
    \caption{Mean retention gap values, computed on the evaluation set, for Jeffreys distance with varying values of $K$ and covariance shapes.}
    \label{figsup:methods-evaluation}
\end{figure}
\newpage

\subsection{Pairwise divergence visualisation with calibrated parameters}
\begin{figure}[ht]
    \centering
    \includegraphics[width=0.7\linewidth]{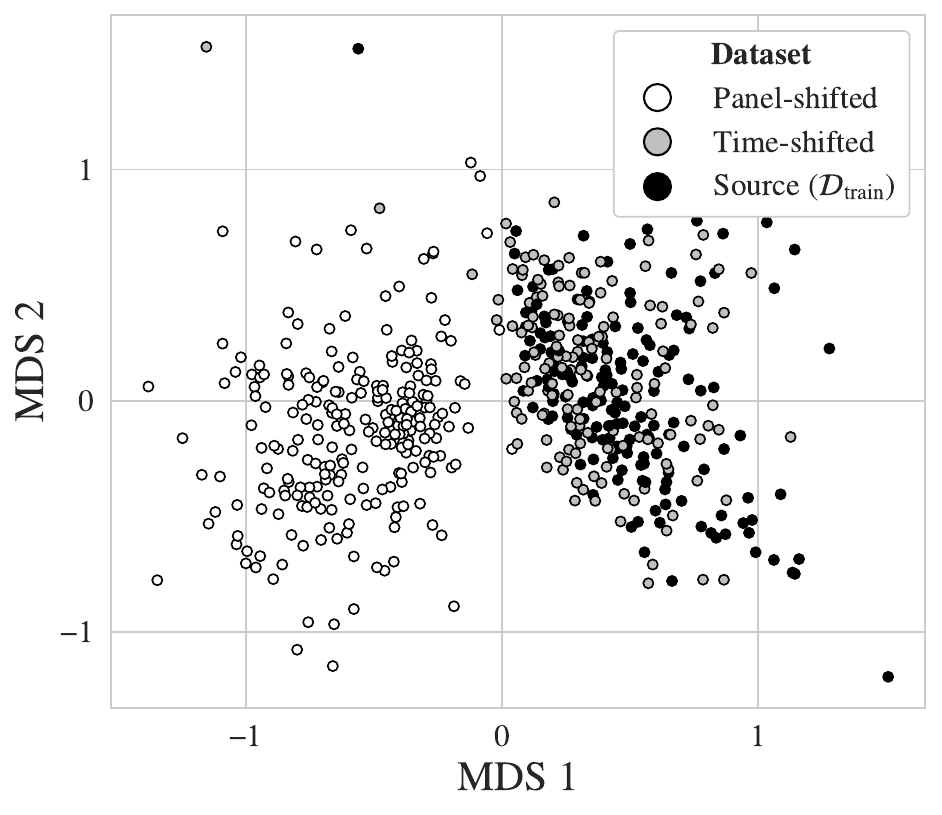}
    \caption{Visualisation of inter-cohort divergence as described in Section~\ref{sec:setup-pairwise},
    using the optimal GMM configuration for the calibration set (Section~\ref{sec:setup-rejection}, $K=1$, full covariance).
    A PERMANOVA confirmed association between cohort membership and the divergence structure ($P=10^{-3}, R^2=0.63$).}
    \label{figsup:mds-k-1}
\end{figure}
\newpage

%
%
%
\bibliographystyle{splncs04}
\bibliography{RECOMB}

\end{document}